\documentclass[conference]{IEEEtran}
\IEEEoverridecommandlockouts
\usepackage{cite}
\usepackage{amsmath,amssymb,amsfonts}
\usepackage{algorithmic}
\usepackage{graphicx}
\usepackage{multicol}
\usepackage{tabularx}
\usepackage{steinmetz}
\usepackage{mathtools}
\usepackage{booktabs}
\usepackage{textcomp}
\usepackage{xcolor}
\usepackage{wrapfig}
\usepackage{subfig}
\usepackage{pifont}

\usepackage[colorlinks,
            linkcolor=blue,
            anchorcolor=blue,
            citecolor=blue]{hyperref}

\def\BibTeX{{\rm B\kern-.05em{\sc i\kern-.025em b}\kern-.08em
    T\kern-.1667em\lower.7ex\hbox{E}\kern-.125emX}}
\begin{document}
\title{Binary Complex Neural Network Acceleration on FPGA \\
{\LARGE (Invited Paper)} 
}

\author{\IEEEauthorblockN{ \textsuperscript{$\star$}Hongwu Peng\textsuperscript{[1]}, \textsuperscript{$\star$}Shanglin Zhou\textsuperscript{[1]}, Scott Weitze\textsuperscript{[2]}, Jiaxin Li\textsuperscript{[1]}, Sahidul Islam\textsuperscript{[3]}, Tong Geng\textsuperscript{[4]}, Ang Li\textsuperscript{[4]}, \\ Wei Zhang\textsuperscript{[1]}, Minghu Song\textsuperscript{[1]}, Mimi Xie \textsuperscript{[3]}, Hang Liu\textsuperscript{[2]}, and Caiwen Ding\textsuperscript{[1]}}
\IEEEauthorblockA{
\textsuperscript{$\star$}These authors contributed equally. \\
\textsuperscript{[1]}University of Connecticut, Storrs, CT, USA. 
\textsuperscript{[2]}Stevens Institute of Technology, Hoboken, NJ, USA. \\
\textsuperscript{[3]}University of Texas at San Antonio, San Antonio, TX, USA. 
\textsuperscript{[4]}Pacific Northwest National Laboratory, Richland, WA, USA. \\
\textsuperscript{[1]}\{hongwu.peng, shanglin.zhou, jiaxin.3.li\}@uconn.edu, \textsuperscript{[1]}wei.zhang.gbs@gmail.com, \\  \textsuperscript{[1]}\{minghu.song, caiwen.ding\}@uconn.edu,  \textsuperscript{[2]}\{sweitze, hliu77\}@stevens.edu,  \\
\textsuperscript{[3]}\{sahidul.islam, mimi.xie\}@utsa.edu, \textsuperscript{[4]}\{tong.geng, ang.li\}@pnnl.gov 
}
}

\maketitle

\begin{abstract}
Being able to learn from complex data with phase information is imperative for many signal processing applications. Today’s real-valued deep neural networks (DNNs) have shown efficiency in latent information analysis but fall short when applied to the complex domain. Deep complex networks (DCN) , in contrast, can learn from complex data, but have high computational costs; therefore, they cannot satisfy the instant decision-making requirements of many deployable systems dealing with short observations or short signal bursts. Recent, Binarized Complex Neural Network (BCNN), which integrates DCNs with binarized neural networks (BNN), shows great potential in classifying complex data in real-time. In this paper, we propose a structural pruning based accelerator of BCNN, which is able to provide more than 5000 frames/s inference throughput on edge devices. The high performance comes from both the algorithm and hardware sides. On the algorithm side, we conduct structural pruning to the original BCNN models and obtain 20 $\times$ pruning rates with negligible accuracy loss; on the hardware side, we propose a novel 2D convolution operation accelerator for the binary complex neural network. Experimental results show that the proposed design works with over 90\% utilization and is able to achieve the inference throughput of 5882 frames/s and 4938 frames/s for complex NIN-Net and ResNet-18 using CIFAR-10 dataset and Alveo U280 Board.
\end{abstract}

\begin{IEEEkeywords}
Binarized neural networks, binarized Complex Neural Network, FPGA, high level synthesis, convolutional neural network, surrogate Lagrangian relaxation
\end{IEEEkeywords}

\section{Introduction}


Due to the growing need for DNN performance on different tasks, today's DNN model has a relatively large model parameter size. For example, ResNet-18 for image classification has a model size of 45 MB, the YOLOv5l of the YOLOv5 family~\cite{YOLOv5} for object detection has a model size as 245 MB, BERT~\cite{devlin2018bert} for NLP has 23M embedding weights and 317M neural network weights. For the shallow DNN model, all the weight and intermediate activations can be stored in the FPGA on-chip memory~\cite{shi2020ftdl} (BRAM and URAM). However, large DNN models' weight is hard to store in the FPGA on-chip memory, and external memory is used to store the weight and activations.~\cite{li2020ftrans}. Therefore, the external memory usage for weight and activation will stall the system's performance.

Using Xilinx Alveo U200 Accelerator Cards~\cite{u200}, as an illustration, we often need to access off-chip memory since the on-chip memory capacity is only 35 MB. However, the off-chip memory is inherently slower to access than on-chip memory in terms of 2 aspects: \textbf{\ding{172}} The off-chip memory (DDR4) total bandwidth is 77 GB/s, while the total bandwidth on-chip memory is 31 TB/s (400$\times$ higher than the off-chip memory); \textbf{\ding{173}} The off-chip memory (DDR4) has around 100 ns memory access latency~\cite{choi2021hbm}, while the on-chip memory can be access at 500 MHz frequency~\cite{shi2020ftdl} with only one cycle latency (50$\times$ lower latency than off-chip memory). Furthermore, The DRAM memory usually has more than 100$\times$ higher power consumption \cite{horowitz2014energy} than the FPGA on-chip SRAM. The whole system performance is constrained by accessing off-chip memory.

In order to fit the model into the FPGA platforms' on-chip memory, model compression techniques can be used. The model compression techniques help reduce the model size and accelerate the inference of the DNN model with an acceptable accuracy loss. The model compression techniques can be divided into two types: weight pruning and weight quantization. We will leverage both of those two techniques in this paper to achieve a higher model compression rate. 

In this paper, we focus on the acceleration of BCNN based NIN-Net and ResNet models. We further propose hardware design to achieve high parallelism and high throughput for the FPGA platform. We implement the proposed techniques on Alveo U280 hardware platforms for comparison of latency and throughput. Experimental results show that the FPGA hardware design enables 5882 frames/s and 6154 frames/s for BCNN based NIN-Net and ResNet models. Our contributions are summarized as follows:

\begin{itemize}
\item Basic framework and building blocks for BCNN are given. For the pooling layer, the spectral pooling, average pooling, and max pooling are compared in terms of model performance. 
\item The Surrogate Lagrangian Relaxation (SLR) weight pruning technique is adopted for BCNN weight pruning, straight-through estimator (STE) is adopted for BCNN weight quantization, the whole model compression framework achieves a high compression ratio with an acceptable accuracy loss. 
\item The binarized complex convolution kernel design is proposed to enable a high level of hardware parallelism and low pipeline initiation interval. 
\item The hardware resource scheduling for the BCNN model implementation on FPGA is discussed, and we achieve a high overall hardware throughput. 
\end{itemize}

The organization of the work is as follows. Section~\ref{Related_work} gives the basics of DNN model compression knowledge and the BCNN model. Section~\ref{BCNN_training} discusses the model structure, SLR-based compression, and STE-based binary quantization. Section~\ref{hardware} gives the hardware design for the BCNN based models. Section~\ref{Experiment} gives the BCNN model's training, the hardware implementation result. Section~\ref{Conclusion} gives the overall conclusion for the hardware design and experiments. 

\section{Related Work}\label{Related_work}

In this section, we will briefly discuss the current works on DNN model compression techniques, BNNs, and complex neural networks.

\subsection{Model Compression}\label{Compression}

In order to reduce the DNN model size and inference latency, the model compression techniques can be adopted. The main challenge of the SOTA model compression techniques is maintaining the model's accuracy on tasks while improving the model inference speed and throughput on hardware platforms. There are two types of model compression techniques: weight pruning \cite{li2020ftrans} and weight quantization \cite{ding2017circnn}. 

Two types of pruning methods are widely used for the weight pruning technique: the unstructured pruning method and the structured pruning method \cite{peng2021accelerating}. The unstructured pruning technique \cite{cai2020yolobile, li2020efficient, yuan2019ultra} is easier to implement on software and has a high compression ratio with low accuracy loss. However, due to its irregular memory access pattern, the unstructured pruning can hardly be accelerated on most of the hardware platforms. The structured pruning technique \cite{cai2020yolobile, li2020efficient, yuan2019ultra} constrains the weight matrix to be pruned in a structured and hardware-friendly pattern. For example, block-circulant matrices \cite{ding2017circnn, lu2017evaluating, liao2017energy} can be used for weight representation after pruning. The structured pruning-based hardware implementation achieves better performance due to the higher parallelism achievable by regular memory access patterns and reduced computation burden. 

Another source of redundancy in the DNN model is the bit representation of the weight. The DNN baseline model usually adopts float32 bits representation for the weight value. In order to compress the bit representation of the data, various works \cite{lin2015neural, wu2016quantized} have proposed different DNN model quantization techniques, including fixed bit-length, ternary, and binary weight representations. The truncated length bit representation reduces DNN model size, computation burden on the hardware platform, and memory bandwidth consumption. The fixed bit-length representation of the DNN model parameter can be further classified into equal-distance quantization and power-of-two quantization. The equal-distance quantization is similar to fixed-point number representation, and it reduces the hardware resource utilization while maintaining high accuracy. The power-of-two quantization further improves the hardware efficiency owning to the bit shift-based multiplication. However, the unequally distributed scale of power-of-two quantization leads to a non-negligible DNN model accuracy degradation. In order to improve the model accuracy and maintain hardware efficiency, mixed powers-of-two-based quantization is proposed \cite{ding2019req}. The mixed powers-of-two-based quantization features its' combination of primary powers-of-two and a secondary powers-of-two part, and the multiplier requires a 2-bit shifter and one adder.

\begin{figure}[t]
    \centering
    \centering
\begin{multicols}{2}
\subfloat  [\label{fig:CNN} Original CNN architecture ]  {\includegraphics[width=1\linewidth]{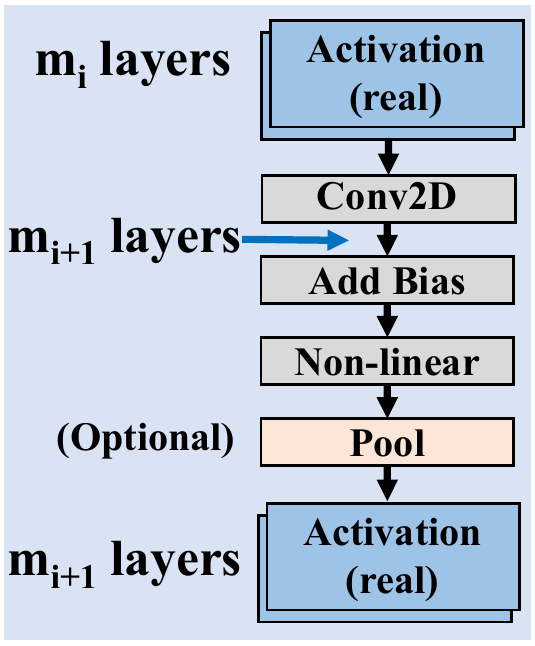}\par }
\subfloat  [\label{fig:BCNN} BCNN architecture]  {\includegraphics[width=1\linewidth]{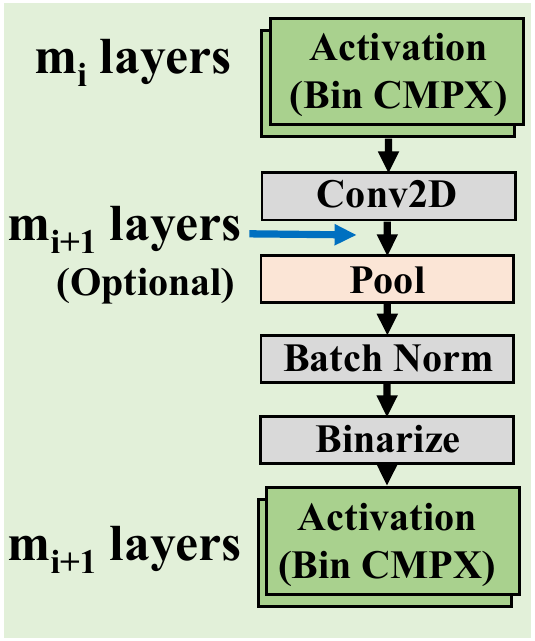}\par }

\end{multicols}

    \caption{Comparison between original CNN and BCNN}
    \label{fig:BNN_CNN}
\end{figure}

\subsection{Binary Complex Neural Networks}\label{BCNN}


The concept of Binary Neural Network (BNN) originated from the binary weight neural network (BWNN) \cite{courbariaux2015binaryconnect}, and the BWNN only quantizes the bit representation of the weight value into the binary value. However, for the FPGA devices with small on-chip memory, the intermediate activations of the BWNN are still too large to be stored in the on-chip SRAM, and external memory is required. The later works \cite{hubara2016binarized, courbariaux2016binarized} proposed BNN and researched the quantization of both activations and weights into a binary representation. Those works illustrated few key concepts to maintain the model accuracy for BNN: straight-through estimator (STE) for gradient descent, batch-normalization after binarized convolution, and keep full precision for both activations and weights at first and last layers. By quantizing both activation and weight, the multiplication is degraded into binary XOR operation and is highly hardware-friendly for FPGA and GPU platforms. A simply popcnt(xor()) function can be used for binarized convolution layers or fully connected layers. 


In order to enhance the model information representation with the same parameter size or even less parameter size, deep, complex neural networks are proposed \cite{trabelsi1705deep}. The complex neural network has the dedicated complex version of the basic building block: convolution, batch normalization, weight initialization strategy, etc. The deep complex neural networks achieve comparable performance to ordinary DNNs with half model parameters. 

The binary complex neural network (BCNN) \cite{li2021bcnn} combines the benefit of both binary neural networks and complex neural networks. The activations and weights of deep complex neural networks are quantized to one bit except the first and last layer. In order to reduce the computation overhead and improve the model accuracy, Li \textit{et al.}. \cite{li2021bcnn} also proposed few new concepts: quadrant binarization for forward and backward propagation, complex Gaussian batch normalization (CGBN), and binary complex weight initialization. Those concepts will be discussed in detail in Section~\ref{BCNN_training}.

\subsection{Basic of Surrogate Lagrangian Relaxation (SLR)}\label{SLR}

The surrogate Lagrangian relaxation method (SLR) \cite{bragin2015convergence} is an optimization algorithm similar to the alternating direction method of multipliers (ADMM) \cite{boyd2011distributed}, which breaks optimization problems into several smaller subproblems that can be solved iteratively. However, it also overcame major convergence difficulties of standard Lagrangian relaxation. As the solution of decomposed subproblems is coordinated by updating Lagrangian multipliers, convergence can be proved when the solutions satisfy the ``surrogate" optimality condition with a novel step-sizing formula \cite{bragin2015convergence}.

Gurevin \textit{et al.} \cite{gurevin2020enabling} was the first work implementing an SLR-based framework on DNN weight pruning. Comparing with the ADMM-based weight pruning method, when under classification task and the same compression rate, SLR can achieve almost $10\%$ point higher accuracy on VGG-16 on CIFAR-10 dataset, and $3\%$ point higher accuracy on ResNet-18 on ImageNet dataset. Under object detection tasks on COCO 2014 benchmark, models pruned with the SLR method can at most have $9\%$ higher accuracy after hard-pruning than the ADMM method under all the YOLO framework. Also, when hardpruning accuracy is checked periodically during training steps, SLR can faster converge and reach the desired accuracy. Almost $3\times$ faster can be achieved during classification tasks for SLR on CIFAR-10 and $2\times$ faster on ImageNet. During objection detection tasks, $3\times$ faster can be achieved on COCO 2014 benchmark. Experiments also show that the SLR-based weight-pruning optimization approach achieves high accuracy even at the hard-pruning stage. This retrain-free propriety reduces the traditional three-stage pruning pipeline to two-stage and reduces the budget of retraining epochs.

\section{Training and Compression on BCNN}\label{BCNN_training}

In this section, we will discuss the BCNN model in detail, which includes model structure, fundamental building blocks and operations for BCNN, weight pruning using SLR, and weight quantization based on quadrant binarization and STE.

\subsection{Structure of BCNN}\label{BCNN_structure}

The comparison between BCNN and the original convolution neural network (CNN) structure is demonstrated in Fig.~\ref{fig:BNN_CNN}. The mathematical equation for convolution and add bias operation can be found in Eq.~\ref{eq:Conv}. As shown in Fig.~\ref{fig:CNN}, the original CNN is composed of convolution layer, add bias, non-linear layer, and pooling layer. For the BCNN, the structure is different from the original CNN, and the structure is shown in Fig.~\ref{fig:BCNN}. The pooling layer and batch normalization layer should come after the convolution layer, and the bias can be removed from the network to reduce the computation overhead without accuracy loss. For the BCNN, batch normalization is a mandatory operation for model convergence \cite{anderson2017high}. 
\begin{equation}
x_{m_{i+1}} = f(\sum_{c=1}^{m_{i}} x_c * w_{c, m_{i+1}} + b_{m_{i+1}})
\label{eq:Conv}
\end{equation}

\begin{figure}[t]
\centering
\includegraphics[width=.45\textwidth]{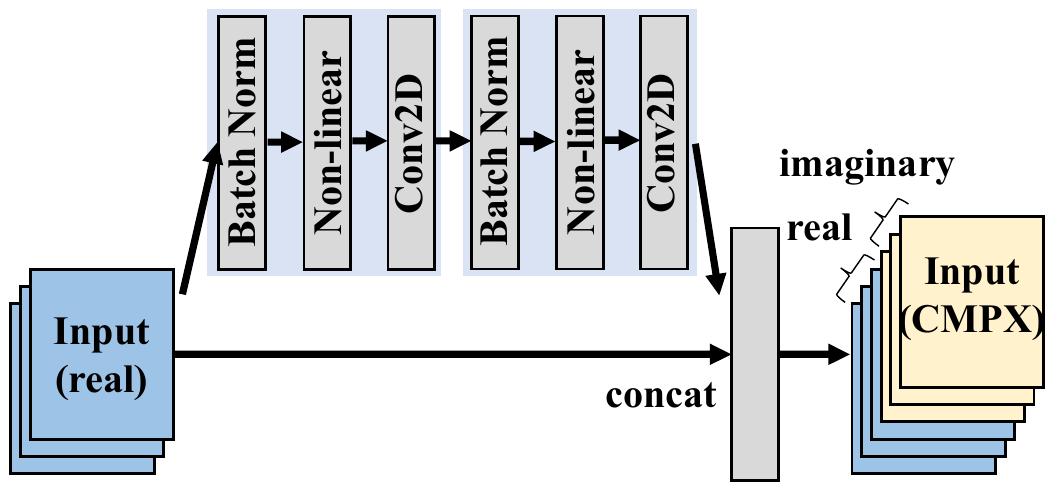}
\caption{Complex input generation}
\label{fig:Complex_gene}
\end{figure}

For the image with three channels (RGB) as input, the initial input only contains real part. In order the generate the complex input, a two-layers residual CNN is designed to learn the imaginary part. The network for generating the complex input is shown in Fig.~\ref{fig:Complex_gene}.

\subsection{Building Blocks and Operations}\label{BBO}

The basic building blocks of BCNN are slightly different from that of ordinary DNN. The complex version of the convolution layer, pooling layer, batch normalization, and binarize function will be discussed in this section. 

\subsubsection{Complex Convolution}

The binary complex number can be defined as: $x_c = x_r + i \cdot x_i$, where the numbers $x_r$ and $x_i$ belong to $\{+1, -1\}$. A single binary complex number is represented by 2 digits, and has twice memory occupations than that of binary number. Assuming the input activation is  $x_c = x_r + i \cdot x_i$, weight is $w_c = w_r + i \cdot w_i$. The dot product between the input activation $x_c$ and weight $w_c$ can be donated as equation format Eq.~\ref{eq:complex_mul} or matrix format Eq.~\ref{eq:matrix_complex}. The mathematically expression of complex convolution operation can be deduced from Eq.~\ref{eq:Conv} and Eq.~\ref{eq:matrix_complex}. The convolution can be divided into bit-wise XOR operation, popcnt operation, and fixed-point number add/subtract for the binary complex convolution operation. Comparing to full precision convolution operation on the FPGA platform, only a few LUT resources will be used to calculate the binary complex convolution operation, and the memory bandwidth requirement is reduced by 32 $\times$. 
\begin{equation}
y_c = y_r + i \cdot y_i  = (x_r w_r - x_i w_i) + i \cdot (x_r w_i - x_i w_r)
\label{eq:complex_mul}
\end{equation}
\begin{equation}
  \begin{bmatrix}
    y_r \\
    y_i \\
  \end{bmatrix}
  =
  \begin{bmatrix}
    w_r & -w_i \\
    w_i & w_r \\
  \end{bmatrix}
  \cdot
  \begin{bmatrix}
    x_r \\
    x_i \\
  \end{bmatrix} 
  \label{eq:matrix_complex}
\end{equation}

\begin{figure}[t]
\centering
\includegraphics[width=.48\textwidth]{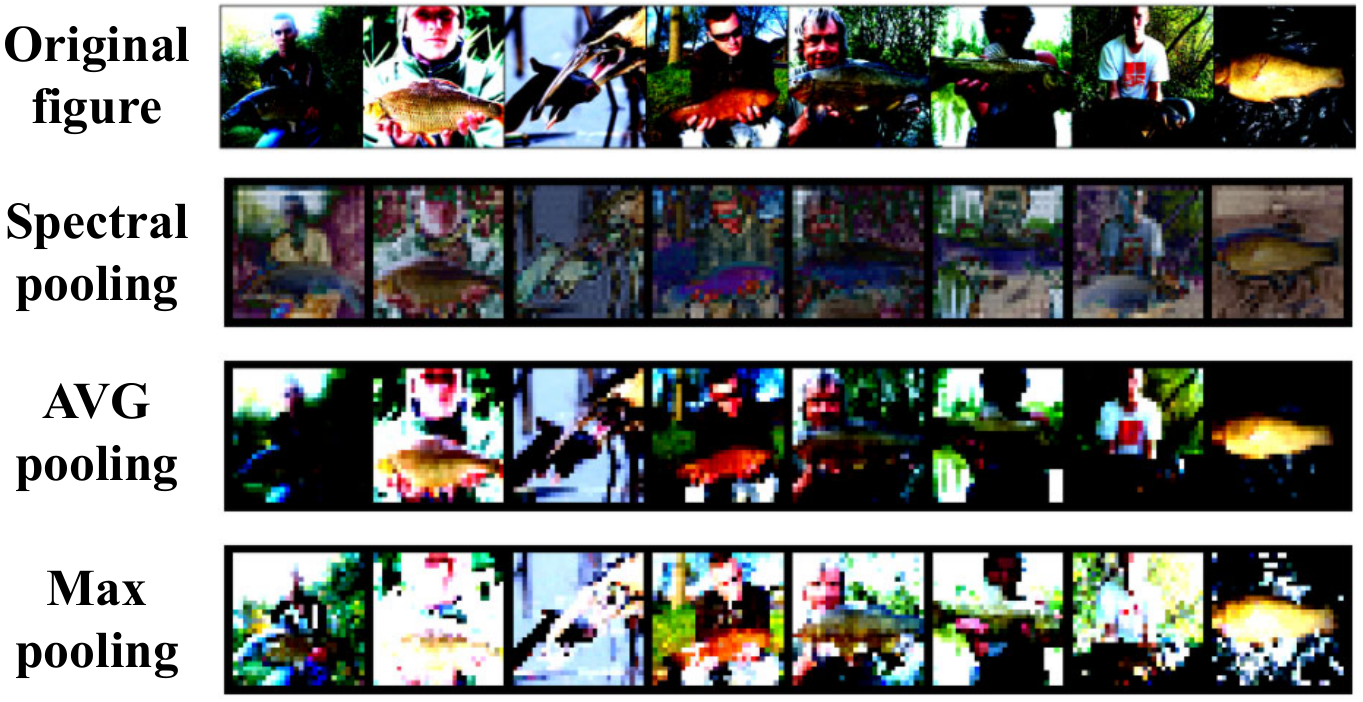}
\caption{Comparison of pooling function}
\label{fig:Comp_fft}
\end{figure}

\subsubsection{Pooling Layer}

The pooling layer is optional in the convolutional layer. For a deep CNN, the pooling layer will be used at specific layers to downsample at a rate to reduce the activation sizes. There are two widely used pooling layers in the original CNN: max pooling and average pool. For BCNN, the activations are represented as complex numbers, enabling the possibility of another type of pooling method that has good preservation of information: spectral pooling \cite{rippel2015spectral}. 

The spectral pooling conducted a fast Fourier transform (FFT) over the 2D dimension of activations and truncated the high-frequency component to leave the center low frequency spectral. An inverse FFT (IFFT) is then conducted to transform the spectral information back to the spatial domain. 

A comparison of different pooling methods is shown in Fig.~\ref{fig:Comp_fft}. The pictures are obtained from the ImageNet dataset \cite{deng2009imagenet}. For spectral pooling, the pixel value is scaled back to the range of $(0, 1)$ to ensure visibility. As shown in the figure, the spectral pooling and average pooling preserves more spatial information than the max pooling. The max pooling is more like a 2D whitening function and performs poorly for images with higher brightness. 

As for the computation complexity, the FFT operation of spectral pooling has a $O(MN\sqrt{MN})$ computation complexity; average pooling and max pooling both has a $O(MN)$ computation. In order to maintain a good trade-off between the model accuracy and computation complexity, the average pooling is chosen for the hardware design.

\subsubsection{CGBN}

The complex version of batch normalization (CGBN) is an essential operation that helps the BCNN to converge. The original form of batch normalization can be found in Eq.~\ref{eq:batch_norm}, where $x_i\textsuperscript{(k)}$ donates the input activation, $\mu_B\textsuperscript{(k)}$ donates the mean of the activations in the mini-batch, $\delta_B\textsuperscript{(k)}$ donates the variance for the activations in the mini-batch, and $\epsilon$ donates a small value to be added to avoid dividing by zero. $\gamma$ and $\beta$ are both scaling factors that can be learned during the training step. The activation output is represented as $\widehat{x}_i\textsuperscript{(k)}$. 
\begin{equation}
  \widehat{x}_i\textsuperscript{(k)} = \gamma \frac{x_i\textsuperscript{(k)} - \mu_B\textsuperscript{(k)} }{\sqrt{(\delta_B\textsuperscript{(k)})\textsuperscript{2} + \epsilon}} + \beta
  \label{eq:batch_norm}
\end{equation}

For the complex neural network, the original form of complex batch normalization can be found in Eq.~\ref{eq:Complex_batch_norm} \cite{trabelsi1705deep}.  $V$ is the $2 \times 2$ covariance matrix, and $E[x_c]$ is the mean of complex activations within the mini-batch. $\gamma$ is a $2 \times 2$ matrix and $\beta$ is a complex number. Both $\gamma$ and $\beta$ can be learned during backpropagation. 
\begin{equation}
  \widehat{x}_c = \gamma \cdot V\textsuperscript{-1/2} (x_c-E[x_c]) + \beta
  \label{eq:Complex_batch_norm}
\end{equation}

However, as discussed in Li \textit{et al.}. \cite{li2021bcnn}, the original form of complex batch normalization requires too much calculation, and the CGBN concept is proposed. The mathematical equation for the CGBN can be found in Eq.~\ref{eq:CGBN}. Both  $\gamma$ and $\beta$ are complex numbers scaling factors and can be learned. The CGBN for BCNN has higher accuracy and lower computation complexity so that CGBN will be used for both the software and hardware implementation. 
\begin{equation}
  \widehat{x}_c = \gamma \cdot(\frac{x_r - \mu_r}{\sqrt{2\delta_r\textsuperscript{2} + \epsilon}} - \frac{x_i - \mu_i}{\sqrt{2\delta_i\textsuperscript{2} + \epsilon}}) + \beta
  \label{eq:CGBN}
\end{equation}

\subsubsection{Binarization}

There are two types of widely used binarization \cite{courbariaux2015binaryconnect}: deterministic binarization and stochastic binarization. The equation for deterministic binarization is given in Eq.~\ref{eq:DeBi}, activation is binarized to +1 and -1 according to their sign. The equation for stochastic binarization is given in Eq.~\ref{eq:StBi}. In the equation, the $\delta(x)$ is a hard clipping function and satisfies $\delta(x) = max(0, min(1, \frac{x+1}{2}))$. 
\begin{equation}
  x_{b} = sign(x) = 
    \begin{cases}
      +1 & \text{if $x \geq 0$ }\\
      -1 & \text{if $x < 0$}\\
    \end{cases}       
    \label{eq:DeBi}
\end{equation}
\begin{equation}
  x_{b} = sign(x) = 
    \begin{cases}
      +1 & \text{has probability $p = \delta(x)$}\\
      -1 & \text{has probability $1 - p$}\\
    \end{cases}   
    \label{eq:StBi}
\end{equation}

The stochastic binarization has a better model accuracy performance than the deterministic binarization. However, the implementation of stochastic binarization requires a  stochastic number generator and is expensive for hardware design. So the deterministic binarization will be used for both software and hardware experiment. 

For the complex number with real and imaginary parts, quadrant binarization is proposed in \cite{li2021bcnn} to conduct the binarization and will be used in this paper. The concept of quadrant binarization is naive and straightforward: the real and imaginary part is binarized individually during the forward propagation. 

\subsection{Channel-wise Weight Pruning Using SLR}\label{SLR2}

Consider an $N$-layer DNN indexed as $i \in 1, ..., N$. The collection of weights at each convolutional layer is denoted by $\mathbf{W}_i$ and the collection of corresponding biases is denoted by $\mathbf{b}_i$, and loss function is denoted by $f(\{\mathbf{W}\}_{i=1}^N, \{\mathbf{b}\}_{i=1}^N)$. The objective of channel-wise weight pruning can be done by minimizing the loss function and make it subject to constraints that the number of nonzero channels of the weight in each layer is less than or equal to a predefined number. This can be formulated as Eq.~\ref{loss}.
\begin{equation}
\label{loss}
\begin{array}{ll}
\underset{\left\{\mathbf{W}_{i}\right\},\left\{\mathbf{b}_{i}\right\}}{\operatorname{minimize}} & f\left(\left\{\mathbf{W}_{i}\right\},\left\{\mathbf{b}_{i}\right\}\right), \\
\text { s.t. } & \mathbf{W}_{i} \in \mathbf{C}_{i}, i=1, \ldots, N,
\end{array}
\end{equation}
where $\mathbf{C}_i \coloneqq \{\mathbf{W}_i|$ the number of channels that being zero in $\mathbf{W}_i$ is greater than $\gamma_i\}$, where $\{\gamma_i\}_{i=1}^N$ is a set of predefined hyper-parameter. This can further be equivalently rewritten in an unconstrained form as Eq.~\ref{loss_rew}.
\begin{equation}
\label{loss_rew}
\underset{\left\{\mathbf{W}_{i}\right\},\left\{\mathbf{b}_{i}\right\}}{\operatorname{minimize}} f\left(\left\{\mathbf{W}_{i}\right\},\left\{\mathbf{b}_{i}\right\}\right)+\sum_{i=1}^{N} h_{i}\left(\mathbf{W}_{i}\right)
\end{equation}
The first term represents the nonlinear loss function, and the second represents the non-differentiable penalty term \cite{zhang2018systematic} that $h_i(.)$ is the indicator function of $\mathbf{C}_i$ shown in Eq.~\ref{gx}.
\begin{equation}
\label{gx}
h_{i}\left(\mathbf{W}_{i}\right)=\left\{\begin{array}{ll}
0 & \text { if } \mathbf{W}_i \in \mathbf{C}_i, \: i=1, \ldots, N \\
+\infty & \text { otherwise }
\end{array}\right.
\end{equation}
The problem cannot be solved only analytically or only using stochastic gradient descent. In this case, duplicate variables are introduced as Eq.~\ref{slr_loss}.
\begin{equation}
\label{slr_loss}
\begin{array}{ll}
\underset{\left\{\mathbf{W}_{i}\right\},\left\{\mathbf{b}_{i}\right\}}{\operatorname{minimize}} & f\left(\left\{\mathbf{W}_{i}\right\},\left\{\mathbf{b}_{i}\right\}\right)+\sum_{i=1}^{N} h_{i}\left(\mathbf{Z}_{i}\right), \\
\text { subject to } & \mathbf{W}_{i}=\mathbf{Z}_{i}, \quad i=1, \ldots, N
\end{array}
\end{equation}

To solve the problem, SLR leverages the augmented Lagrangian multipliers and penalizes their violations using quadratic penalties, which can be written as Eq.~\ref{slr_uncon}.
\begin{equation}
\label{slr_uncon}
\begin{aligned}
L_{\rho}&\left(\mathbf{W}_{i}, \mathbf{b}_{i}, \mathbf{Z}_{i}, \boldsymbol{\Lambda}_{i}\right)=f\left(\mathbf{W}_{i}, \mathbf{b}_{i}\right)+\sum_{i=1}^{N} h_{i}\left(\mathbf{Z}_{i}\right) \\
&+\sum_{i=1}^{N} \operatorname{tr}\left[\mathbf{\Lambda}_{i}^{T}\left(\mathbf{W}_{i}-\mathbf{Z}_{i}\right)\right]+\sum_{i=1}^{N} \frac{\rho}{2}\left\|\mathbf{W}_{i}-\mathbf{Z}_{i}\right\|_{F}^{2}
\end{aligned}
\end{equation}
where $\Lambda_n$ is the Lagrangian multipliers (dual variables) corresponding to constraints $\mathbf{W}_i=\mathbf{Z}_i$ with the same dimension as $\mathbf{W}_i$. The scalar $\rho$ is a positive number that represents the penalty coefficient, and $\|.\|_F^2$ denotes the Frobenius norm. This can be decomposed into two subproblems and being solved iteratively until convergence.

\textbf{Step 1: Solve subproblem (loss function) for $\mathbf{W}_{i}$ using Stochastic Gradient Decent.} At iteration $k$, the ``loss function" subproblem is minimizing the Lagrangian function while keeping $\mathbf{Z}_{i}$ at previously obtained values $\mathbf{Z}_{i}^{k-1}$ for given values of multipliers $\Lambda_n^k$: $\min _{\mathbf{W}_{i}} L_{\rho}\left(\mathbf{W}_{i}, \mathbf{b}_i, \mathbf{Z}_{i}^{k-1}, \boldsymbol{\Lambda}_{i}\right)$. This can be solved by using stochastic gradient descent (SGD) \cite{bottou2010large}. ``Surrogate" optimality condition (Eq.~\ref{slr_1}) needs to be satisfied to ensure multipliers update to right directions.
\begin{equation}
\label{slr_1}
L_{\rho}\left(\mathbf{W}_{i}^{k}, \mathbf{b}_i^k, \mathbf{Z}_{i}^{k-1}, \mathbf{\Lambda}_{i}^{k}\right) < L_{\rho}\left(\mathbf{W}_{i}^{k-1}, \mathbf{b}_i^{k-1}, \mathbf{Z}_{i}^{k-1}, \mathbf{\Lambda}_{i}^{k}\right)
\end{equation}
When Eq.~\ref{slr_1} is satisfied, stepsizes and multipliers are updated as Eq.~\ref{slr_1_upd}.
\begin{equation}
\label{slr_1_upd}
\begin{aligned}
&s^{\prime k}=\alpha^{k} \frac{s^{k-1}|| \mathbf{W}^{k-1}-\mathbf{Z}^{k-1}||}{\left\|\mathbf{W}^{k}-\mathbf{Z}^{k-1}\right\|} \\
&\mathbf{\Lambda}_{i}^{k+1}:=\mathbf{\Lambda}_{i}^{k}+s^{\prime k}\left(\mathbf{W}_{i}^{k}-\mathbf{Z}_{i}^{k-1}\right)
\end{aligned}
\end{equation}
Otherwise, previous stepsizes and multipliers are kept.

\begin{figure}[t]
\centering
\includegraphics[width=.48\textwidth]{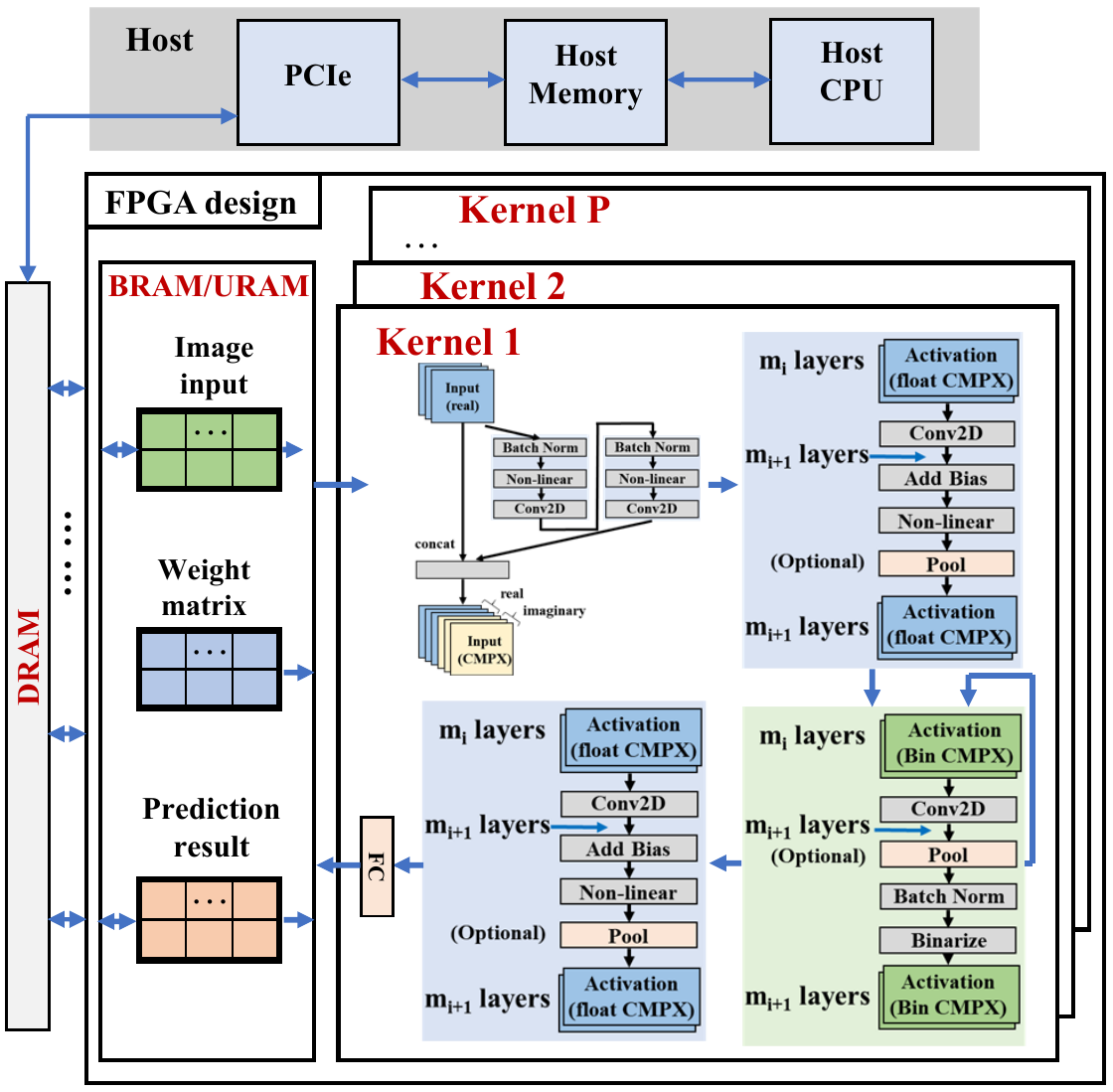}
\caption{Hardware design architecture}
\label{fig:Hd_Archi}
\end{figure}

\textbf{Step 2: Solve subproblem (Channel-wise pruning) for $\mathbf{Z}_{i}$ through Pruning using Projections onto Discrete Subspace.} The channel-wise pruning subproblem can be written as: $\min _{\mathbf{Z}_{i}} L_{\rho}\left(\mathbf{W}_{i}^{k}, \mathbf{b}_{i}^{k},  \mathbf{Z}_{i}, \mathbf{\Lambda}_{i}^{k+1}\right)$. The global optimal of this subproblem can be solved analytically as $h_i(.)$ is the indicator function. For the weight tensor in each convolutional layer, Frobenius norm of each channel of the tensor is calculated and denoted as $\{F\}_c$, where $c=1,2,...,C_i$ and $C_i$ is the number of channels in the tensor. Channels with ``larger" $\{F\}_c$ are kept and with ``smaller" $\{F\}_c$ are set to zeros following that the number of nonzero channels is less than or equal to $\gamma_i$. 
Second ``surrogate" optimality condition (Eq.~\ref{slr_2}) needs to be satisfied to ensure multipliers update to right directions.
\begin{equation}
\label{slr_2}
L_{\rho}\left(\mathbf{W}_{i}^{k}, \mathbf{b}_{i}^{k}, \mathbf{Z}_{i}^{k}, \boldsymbol{\Lambda}_{i}^{\prime k+1}\right) < L_{\rho}\left(\mathbf{W}_{i}^{k}, \mathbf{b}_{i}^{k}, \mathbf{Z}_{i}^{k-1}, \boldsymbol{\Lambda}_{i}^{\prime k+1}\right)
\end{equation}
When Eq.~\ref{slr_2} is satisfied, stepsizes and multipliers are updated again as Eq.~\ref{slr_2_upd}.
\begin{equation}
\label{slr_2_upd}
\begin{aligned}
&s^{k}=\alpha^{k} \frac{s^{\prime k}|| \mathbf{W}^{k-1}-\mathbf{Z}^{k-1} \|}{\left\|\mathbf{W}^{k}-\mathbf{Z}^{k}\right\|} \\
&\mathbf{\Lambda}_{i}^{k+1}:=\mathbf{\Lambda}_{i}^{\prime k+1}+s^{k}\left(\mathbf{W}_{i}^{k}-\mathbf{Z}_{i}^{k}\right)
\end{aligned}
\end{equation}
Same as the first step, previous stepsizes and multipliers are kept if the condition is not satisfied. In both steps, stepsize-setting parameters are formalized as Eq.~\ref{alpha}, where $M$ and $r$ are predefined hyper-parameters.
\begin{equation}
\label{alpha}
\alpha^{k}=1-\frac{1}{M \times k^{\left(1-\frac{1}{k^{r}}\right)}}, \quad M>1,\: 0<r<1
\end{equation}

\subsection{Weight Quantization}\label{STE}

The pruned model from Section~\ref{SLR2} will be used for further weight quantization. The binarization is conducted for both activations and weights in the binarized convolution layer, and deterministic binarization function is used. The binarization function (Eq.~\ref{eq:DeBi}) is non-differentiable at 0, so the direct back-propagation is not feasible for the weight quantizaiton training. Straight-Through-Estimator (STE) is proposed in previous literatures \cite{courbariaux2015binaryconnect, hubara2016binarized} for the back-propagation. The complex version of STE is proposed in \cite{li2021bcnn}, and the equation can be found in Eq.~\ref{eq:STE}. 
\begin{equation}
  \frac{\partial Loss}{\partial w} = \frac{\partial Loss}{\partial w_{rb}} 1_{|w_r|<C_{clip}} + \frac{\partial Loss}{\partial w_{ib}} 1_{|w_i|<C_{clip}}
    \label{eq:STE}
\end{equation}

\section{FPGA Hardware Architecture}\label{hardware}

FPGA is the one of the most popular hardware platform for DNN applications. FPGA platform features it's reconfigurable structure and high level of parallelism  for hardware design. With the growing size of the DNN model, the weight matrices and the activations are too large to be stored in the FPGA on-chip memory. However, the aforementioned pruning and weight quantization technologies compressed both activations and weights representation, making it possible for FPGA platform to store all the intermediate results within on-chip memory. In this section, hardware design will be conducted based on Vivado HLS 2020.1, we will present our FPGA hardware structure for the BCNN model.

\subsection{Overall FPGA structure}\label{overall_structure}

Our overall FPGA structure for BCNN is shown in Fig.~\ref{fig:Hd_Archi}. We'll have 2 BCNN model design for the this section: network in network (NIN) \cite{lin2013network} based BCNN model and ResNet-18 based BCNN model. Both of those 2 models will be composed of 3 major layers: complex input generation layer (Fig.~\ref{fig:Complex_gene}), full precision complex convolutional layer (Fig.~\ref{fig:CNN}), binarized complex convolutional layer (Fig.~\ref{fig:BCNN}). A fully connected (FC) layer will be used at last to generate the prediction output. 

\begin{figure}[ht]
\centering
\includegraphics[width=.48\textwidth]{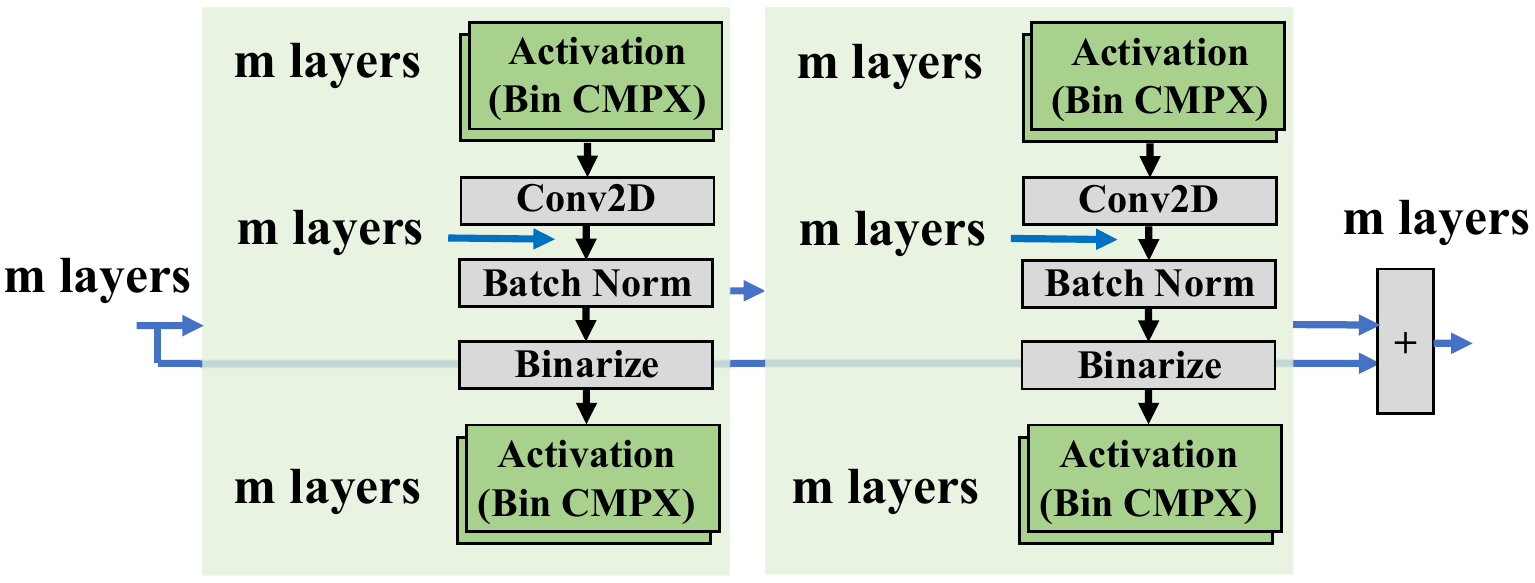}
\caption{Residual blocks 1}
\label{fig:ResB1}
\end{figure}

\begin{figure}[ht]
\centering
\includegraphics[width=.48\textwidth]{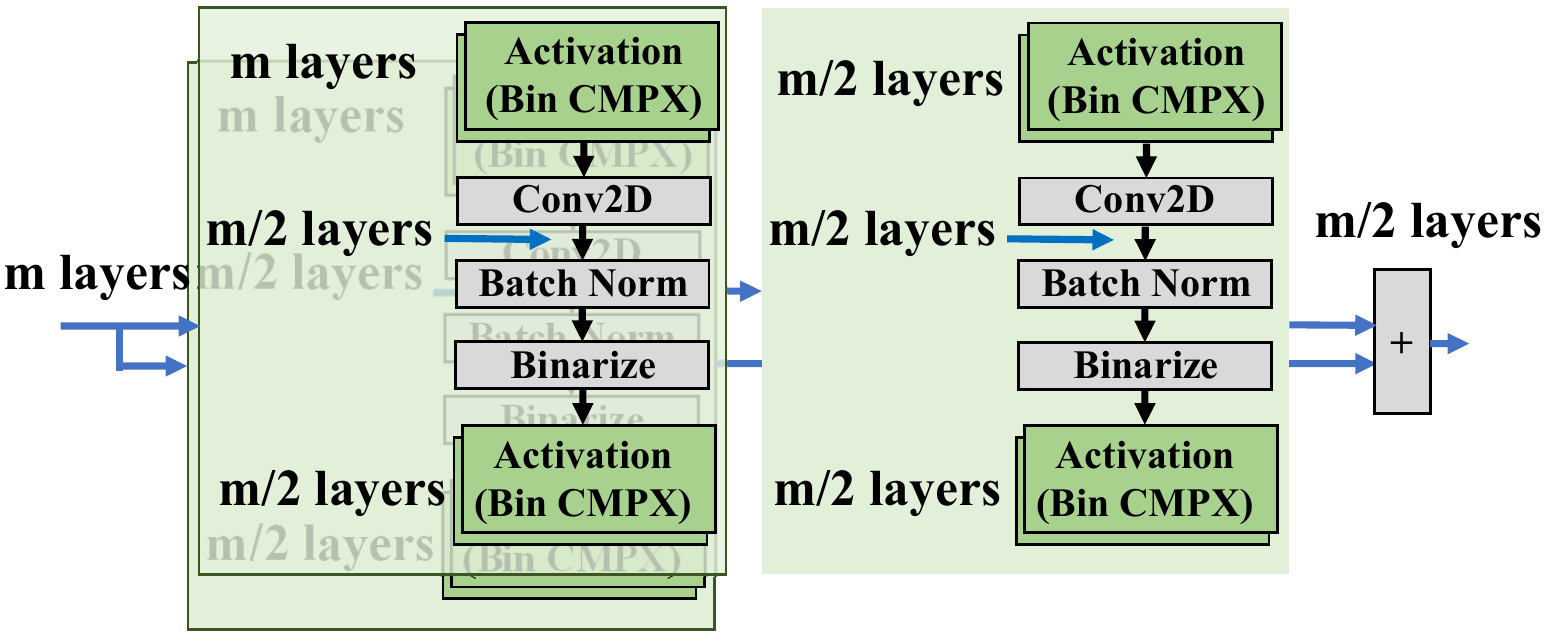}
\caption{Residual blocks 2}
\label{fig:ResB2}
\end{figure}

For the ResNet-18 network, there are 2 types of residual blocks, and both of those residual blocks are binarized block for the BCNN model. The residual block 1 is shown in Fig.~\ref{fig:ResB1}, the input is passed through 2 binarized complex convolutional layers and is added with origin input to get the final output. The residual block 2 is shown in Fig.~\ref{fig:ResB2}, one of the path has 2 binarized complex convolutional layers and another path has only 1 binarized complex convolutional layers, then outputs of those 2 paths are added together to generate the final output.

\subsection{Hardware Design Details}\label{hardware_details}

There are several major building blocks for the hardware design: full precision complex convolutional layer, batch normalization, and pooling layer; binarized complex convolutional layer, batch normalization, and pooling layer. The activation functions used in the models are simple ReLU and Hardtanh functions which have very low computation costs. 

\begin{figure}[t]
\centering
\includegraphics[width=.48\textwidth]{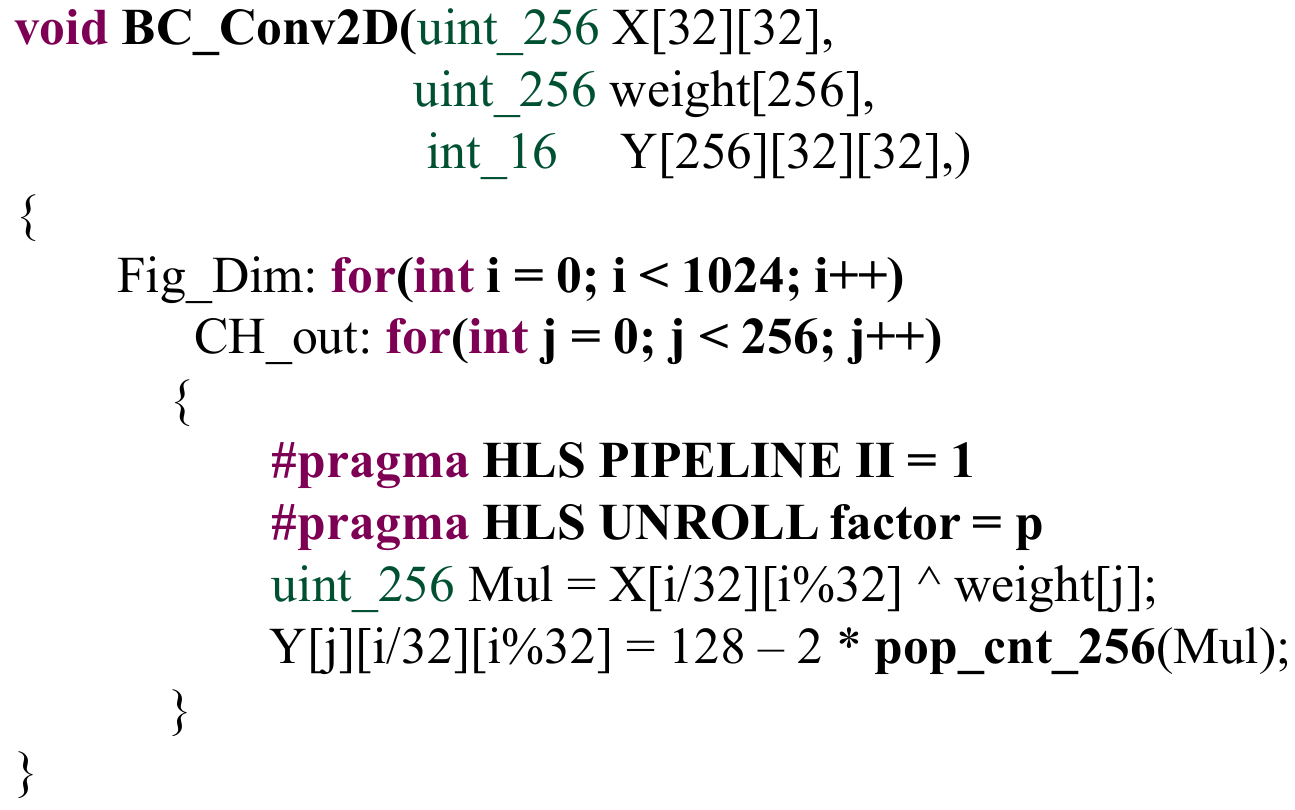}
\caption{Binarized complex convolutional operation}
\label{fig:BC_Conv2d}
\end{figure}

\begin{figure}[ht]
\centering
\includegraphics[width=.48\textwidth]{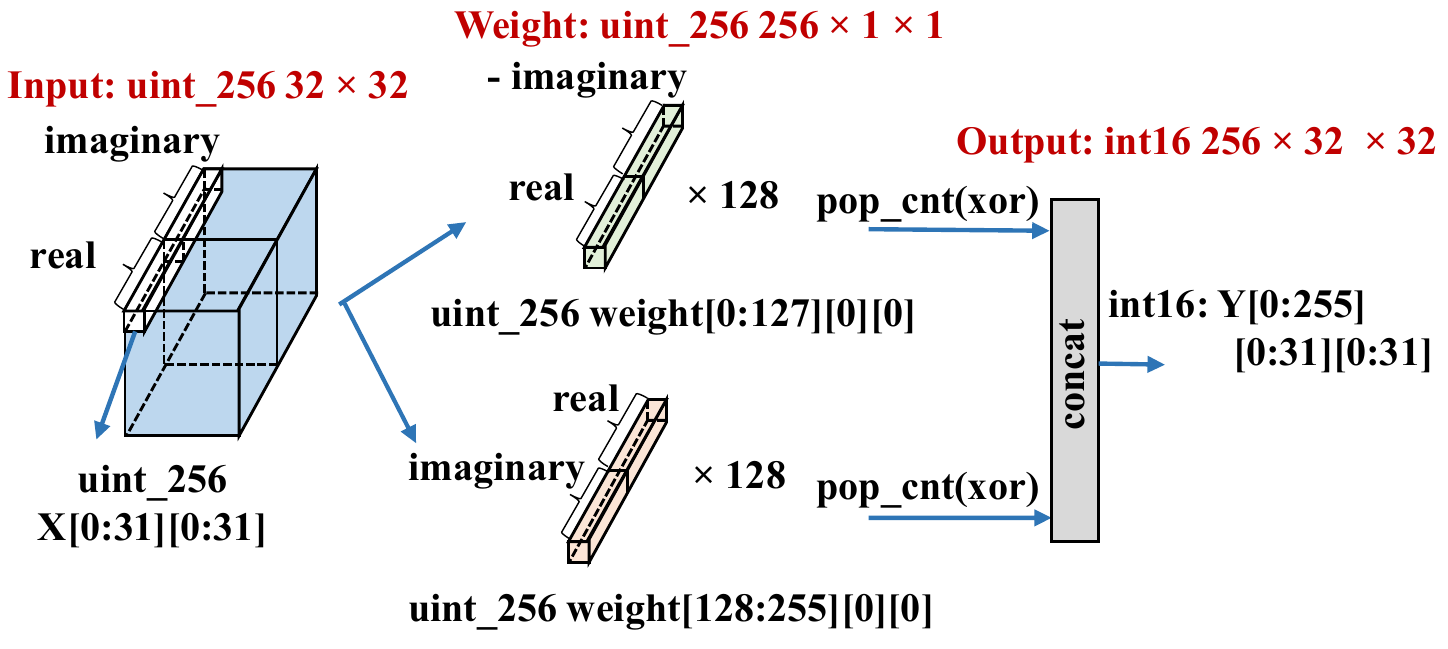}
\caption{Hardware mapping for binarized complex convolutional operation}
\label{fig:BC_Conv2d_map}
\end{figure}

The binarized complex convolution computation follows Eq.~\ref{eq:Conv}, Eq.~\ref{eq:matrix_complex}, and Eq.~\ref{eq:DeBi}. The weight matrix for real and imaginary part can be concatenated together to get the final weight matrix input. The concatenated input can directly serve as the general weight matrix input for the convolution. The example HLS code for the binarized  convolution with is given in Fig.~\ref{fig:BC_Conv2d}, and the hardware mapping is shown in Fig.~\ref{fig:BC_Conv2d_map}. In the example design, the convolution kernel size is $1 \times 1$, stride is 1, and input and output number of complex channel are both 128. For both the input and output channel, the first 128 channels are the real part and the last 128 channels are imaginary parts. The output channel and input channel are chosen for 2 levels of parallelism. The sparse channel pruning for the convolutional layer can be conducted during the binarization operation to avoid pipeline stall  during the convolution operation.

\section{Experiment}\label{Experiment}

\subsection{Training of BCNN Models}

In this section, we will apply SLR pruning and STE-based quantization for both BCNN based NiN-Net and ResNet-18. For the CIFAR-10 dataset, we'll present the training result for both BCNN based NiN-Net and ResNet-18. For ImageNet dataset, we'll only present the result of BCNN based ResNet-18. We conduct our experiments on Ubuntu 18.04, Python 3.7 and PyTorch v1.6.0 software version. And we are using Nvidia Quadro RTX 6000 GPU with 24 GB GPU memory for the training.

Firstly, in order to finalize the pooling layer function for the final model, spectral pooling, average pruning, and max pooling are compared on the BCNN model. BCNN based NIN-Net is used for demonstration. Three types of pooling are compared in terms of their accuracy achievable, and the result is given in Table~\ref{table:pooling_acc}. The average pooling has better performance than the other two types of pooling methods with an acceptable computation complexity. So the average pooling will be used for the BCNN model. 


\begin{table}[htbp]   
    \centering
    \caption{Pooling layer comparison}
    \begin{tabular}{rrr}
    \hline
    \hline
    \multicolumn{1}{c|}{Network} & \multicolumn{1}{c|}{Pooling layer type} & \multicolumn{1}{c}{Accuracy} \\ \hline
    \multicolumn{1}{c|}{} & \multicolumn{1}{c|}{Spectral pooling} & \multicolumn{1}{c}{87.09\%} \\ 
    \multicolumn{1}{c|}{BCNN based NIN-Net} & \multicolumn{1}{c|}{Average pruning} & \multicolumn{1}{c}{87.42\%} \\ 
    \multicolumn{1}{c|}{} & \multicolumn{1}{c|}{Max pooling} & \multicolumn{1}{c}{86.88\%} \\ \hline
    \hline
    \end{tabular}
  \label{table:pooling_acc}
\end{table}

\begin{table}[htbp]   
    \centering
    \caption{Model accuracy for CIFAR-10 dataset }
    \begin{tabular}{rrr}
    \hline
    \hline
    \multicolumn{1}{c|}{Network} & \multicolumn{1}{c|}{Type} & \multicolumn{1}{c}{Accuracy} \\ \hline
    \multicolumn{1}{c|}{ } & \multicolumn{1}{c|}{Original} & \multicolumn{1}{c}{89.13\%} \\ 
    \multicolumn{1}{c|}{NIN-Net} & \multicolumn{1}{c|}{Pruned} & \multicolumn{1}{c}{84.92\%} \\ 
    \multicolumn{1}{c|}{ } & \multicolumn{1}{c|}{Pruned \& quantized} & \multicolumn{1}{c}{83.17\%} \\ \hline
    \multicolumn{1}{c|}{ } & \multicolumn{1}{c|}{Original} & \multicolumn{1}{c}{89.31\%} \\ 
    \multicolumn{1}{c|}{Complex NIN-Net} & \multicolumn{1}{c|}{Pruned} & \multicolumn{1}{c}{86.13\%} \\ 
    \multicolumn{1}{c|}{ } & \multicolumn{1}{c|}{Pruned \& quantized} & \multicolumn{1}{c}{85.12\%} \\ \hline
    \multicolumn{1}{c|}{ } & \multicolumn{1}{c|}{Original} & \multicolumn{1}{c}{92.14\%} \\ 
    \multicolumn{1}{c|}{ResNet-18} & \multicolumn{1}{c|}{Pruned} & \multicolumn{1}{c}{88.51\%} \\ 
    \multicolumn{1}{c|}{ } & \multicolumn{1}{c|}{Pruned \& quantized} & \multicolumn{1}{c}{87.67\%} \\ \hline
    \multicolumn{1}{c|}{ } & \multicolumn{1}{c|}{Original} & \multicolumn{1}{c}{92.51\%} \\ 
    \multicolumn{1}{c|}{Complex ResNet-18} & \multicolumn{1}{c|}{Pruned} & \multicolumn{1}{c}{90.23\%} \\ 
    \multicolumn{1}{c|}{ } & \multicolumn{1}{c|}{Pruned \& quantized} & \multicolumn{1}{c}{89.34\%} \\ \hline
    \hline
    \end{tabular}
  \label{table:Model_acc}
\end{table}
For the complex version of the model, the number of channel is reduced by half to ensure the same model size for the BCNN model. For the weight pruning, the pruning ratio is set as $0.5$ for most of the intermediate layers. Accuracy of four models: NIN-Net, complex NIN-Net, ResNet-18, and complex ResNet-18 on CIFAR-10 dataset on  can be found in Table~\ref{table:Model_acc}.  For the ImageNet dataset, the BCNN based ResNetE-18 model \cite{li2021bcnn} will be used for the pruning and quantization, and the result is given in Table~\ref{table:Model_acc_imagenet}. As shown in the table, the complex version of the network will have better performance than the ordinary counterpart. Thus, the complex binary network-based NIN-Net and ResNet-18 will be used for the hardware design evaluation.


\begin{table}[htbp]   
    \centering
    \caption{Complex ResNetE-18 model accuracy for ImageNet dataset }
    \begin{tabular}{rrrr}
    \hline 
    \hline
    \multicolumn{1}{c|}{Network} & \multicolumn{1}{c|}{Type} & \multicolumn{1}{c}{Top 5 accuracy}\\ \hline
    \multicolumn{1}{c|}{Complex } & \multicolumn{1}{c|}{Original} &  \multicolumn{1}{c}{83.46\%} \\ 
    \multicolumn{1}{c|}{ResNet-18} & \multicolumn{1}{c|}{Pruned} &  \multicolumn{1}{c}{78.38\%} \\ 
    \multicolumn{1}{c|}{ } & \multicolumn{1}{c|}{Pruned \& Quantized} & \multicolumn{1}{c}{71.69\%} \\ \hline
    \hline
    \end{tabular}
  \label{table:Model_acc_imagenet}
\end{table}

\subsection{Hardware Evaluation}

The hardware evaluation is conducted on Xilinx SDSoC 2020.1 and Vivado HLS 2020.1.1. Alveo U290 board is used for the demonstration. The CIFAR-10 and dataset will be used as input, each image input has size of $3 \times 32 \times 32$. The BCNN based NIN-Net model and the BCNN based ResNet-18 model will be design to fit the CIFAR-10 dataset image input. 

The resource utilization for a single BCNN based NIN-Net inference kernel can be found in Table~\ref{table:Rs_NIN}. For a single kernel, the execution latency is 1.53 ms. The maximum resource utilization is bounded by the LUT resources, and nine kernels can be used simultaneously to achieve a higher level of parallelism. The maximum achievable throughput will be 5882 frames/s.

\begin{table}[htbp]   
    \centering
    \caption{Resource utilization for BCNN based NIN-Net}
    \begin{tabular}{rrrr}
    \hline
    \hline
    \multicolumn{1}{c|}{Resource} & \multicolumn{1}{c|}{Utilization} & \multicolumn{1}{c|}{Total} & \multicolumn{1}{c}{percentage (\%)} \\ \hline

    \multicolumn{1}{c|}{DSP} & \multicolumn{1}{c|}{575} & \multicolumn{1}{c|}{9024} & \multicolumn{1}{c}{6.37}\\ \hline

    \multicolumn{1}{c|}{FF} & \multicolumn{1}{c|}{88845} & \multicolumn{1}{c|}{2607360} & \multicolumn{1}{c}{3.41}\\ \hline

    \multicolumn{1}{c|}{LUT} & \multicolumn{1}{c|}{137387} & \multicolumn{1}{c|}{1303680} & \multicolumn{1}{c}{10.54}\\ \hline

    \hline
    \end{tabular}
  \label{table:Rs_NIN}
\end{table}

For the BCNN based ResNet-18 model, the resource utilization for a single inference kernel can be found in Table~\ref{table:Rs_RES}. The execution latency is 1.62 ms.  The resource utilization is also bounded by the LUT resources. In this case, eight kernels can be used simultaneously during the inference step. For the BCNN based ResNet-18, the maximum throughput for the Alveo U280 platform is 4938 frames/s. 

\begin{table}[htbp]   
    \centering
    \caption{Resource utilization for BCNN based ResNet-18}
    \begin{tabular}{rrrr}
    \hline
    \hline
    \multicolumn{1}{c|}{Resource} & \multicolumn{1}{c|}{Utilization} & \multicolumn{1}{c|}{Total} & \multicolumn{1}{c}{percentage (\%)} \\ \hline

    \multicolumn{1}{c|}{DSP} & \multicolumn{1}{c|}{465} & \multicolumn{1}{c|}{9024} & \multicolumn{1}{c}{5.15}\\ \hline

    \multicolumn{1}{c|}{FF} & \multicolumn{1}{c|}{112347} & \multicolumn{1}{c|}{2607360} & \multicolumn{1}{c}{4.31}\\ \hline

    \multicolumn{1}{c|}{LUT} & \multicolumn{1}{c|}{161306} & \multicolumn{1}{c|}{1303680} & \multicolumn{1}{c}{12.37}\\ \hline

    \hline
    \end{tabular}
  \label{table:Rs_RES}
\end{table}

The cross-platform throughput comparison for BCNN model is conducted. The FPGA platform is Alveo U280, and the GPU platform is a single card Nvidia Quadro RTX 6000 GPU with 24 GB GPU memory. The throughput comparison can be found in Table~\ref{table:cross_throughput}. The proposed FPGA design achieved a 1.51 $\times$ speed up on BCNN based NIN-Net model and 1.58 $\times$ speed up on BCNN based ResNet-18 model. 

\begin{table}[htbp]   
    \centering
    \caption{Cross-platform throughput comparison}
    \begin{tabular}{rrrr}
    \hline
    \hline
    \multicolumn{1}{c|}{Model} & \multicolumn{1}{c|}{Platform} & \multicolumn{1}{c}{Throughput (frames/s)}  \\ \hline

    \multicolumn{1}{c|}{BCNN based NIN-Net} & \multicolumn{1}{c|}{Alveo U280} & \multicolumn{1}{c}{5882} \\ 

    \multicolumn{1}{c|}{} & \multicolumn{1}{c|}{RTX 6000} & \multicolumn{1}{c}{3890}\\ \hline

    \multicolumn{1}{c|}{BCNN based ResNet-18} & \multicolumn{1}{c|}{Alveo U280} & \multicolumn{1}{c}{4938} \\ 

    \multicolumn{1}{c|}{} & \multicolumn{1}{c|}{RTX 6000} & \multicolumn{1}{c}{3123}\\ \hline

    \hline
    \end{tabular}
  \label{table:cross_throughput}
\end{table}

\section{Conclusion}\label{Conclusion}

We are the first work to evaluate the BCNN for the FPGA platform. BCNN reduces the memory storage as well as memory bandwidth requirement for the DNN model while maintaining good performance in model accuracy. Further, the resource utilization for BCNN model also reduces since most of the convolution computation will be degraded into XOR and pop\_cnt operations. We utilize the HLS tool to design our BCNN models, and the proposed BCNN model hardware design achieves a 5882 frames/s and 6154 frames/s for BCNN based NIN-Net and BCNN based ResNet-18 on the Alveo U280 FPGA platform, which are 1.51 $\times$ and 1.58 $\times$ speed up than the GPU platform.

{\vspace{\baselineskip}
\bibliographystyle{IEEEtran}
\bibliography{bibligraphy}
}



\end{document}